\documentclass{article}

\PassOptionsToPackage{numbers,compress}{natbib}

\usepackage[preprint]{neurips_2026}
\usepackage{tabularx} 
\usepackage{booktabs} 
\usepackage{caption}
\usepackage{pifont}
\usepackage{multirow}
\newcommand{\cmark}{\ding{51}} 
\newcommand{\xmark}{\ding{55}} 
\newcolumntype{Y}{>{\centering\arraybackslash}X}
\newcolumntype{Z}{>{\raggedright\arraybackslash}X}
\usepackage{graphicx}
\usepackage{subcaption}

\usepackage[utf8]{inputenc} 
\usepackage[T1]{fontenc}    
\usepackage{hyperref}       
\hypersetup{hidelinks}
\usepackage{url}            
\usepackage{booktabs}       
\usepackage{amsmath}        
\usepackage{amsfonts}       
\usepackage{nicefrac}       
\usepackage{microtype}      
\usepackage{xcolor}         
\usepackage{algorithm}
\usepackage{algorithmic}

\title{Radial-Angular Geometry for Reliable Update Diagnosis in Noisy-Label Learning}

\author{%
  Ningkang Peng \\
  Nanjing Normal University
  \And
  Jingyang Mao \\
  Nanjing Normal University
  \And
  Xiaoqian Peng \\
  Nanjing University of Chinese Medicine
  \And
  Weiguang Qu \\
  Nanjing Normal University
  \And
  Yanhui Gu \\
  Nanjing Normal University
}

\begin{document}

\maketitle

\begin{abstract}

Noisy-label methods often estimate sample reliability from forward-space signals such as loss, confidence, or entropy. These signals indicate whether a sample is difficult to predict, but they do not directly test whether its observed label induces a reliable parameter update.
This gap matters because hard clean samples and mislabeled samples can have similar loss while inducing different updates. We recast reliability estimation as diagnosis of the observed-label update. The sample-wise empirical Fisher trace gives a backward-space measure of update energy: for the classifier layer, it factorizes into a prediction-residual term and a feature-sensitivity term, so it captures information beyond scalar loss. Trace, however, is still a radial magnitude signal and cannot decide whether a large update is useful or harmful. We therefore propose Relative Geometric Conflict (RGC), which compares the observed-label gradient with a reference gradient induced by an EMA teacher. The conflict term helps distinguish large but aligned hard-clean updates from large conflicting updates caused by corrupted labels. Across synthetic and real-world noisy-label benchmarks, RGC improves hard-clean preservation and accuracy under our evaluation protocol.

\end{abstract}

\section{Introduction}
Learning with noisy labels (LNL) can be viewed as a problem of unreliable supervision \citep{patrini2017loss,han2018coteaching,li2020dividemix}.
Most existing methods estimate sample reliability from forward-space quantities such as loss, confidence, or entropy \citep{liu2020elr,li2020dividemix}.
This design also appears in more recent sample-selection and label-refinement frameworks such as ProMix and PSSCL \citep{promix,zhang2025psscl}.
These signals are convenient because they are available during training, but they primarily answer how difficult the current prediction is under the observed label.
They leave open the optimization question of whether the update induced by that label is trustworthy.
Hard clean samples and mislabeled samples can therefore overlap in forward space, since both may exhibit high loss or low confidence, especially under instance-dependent or feature-dependent noise \citep{xia2020part,PLC}.

This observation motivates us to shift the reliability criterion from forward difficulty to parameter-space update behavior.
For a sample with observed label $\tilde y_i$, the actual training effect is determined by the gradient
$\mathbf{g}_i^{\mathrm{obs}}=\nabla_\theta \ell(f_\theta(x_i),\tilde y_i)$.
The sample-wise empirical Fisher trace, which reduces to the squared gradient norm for $\mathbf{g}_i\mathbf{g}_i^\top$, gives a backward-space diagnostic of this effect \citep{amari1998natural,kunstner2019limitations}.
For the classifier layer, the trace factorizes as
$\tau_i^{\mathrm{last}}=\|p_i-e_{\tilde y_i}\|_2^2\|h_i\|_2^2$,
showing that it captures not only the prediction residual but also how strongly this residual is transmitted into parameter updates through the feature representation.
Thus, trace provides information beyond the scalar forward loss and can distinguish samples that have similar loss but different update energy.

However, update energy alone is still not reliability.
A hard clean sample may induce a large but useful update, while a mislabeled sample may induce a large and misleading update.
Therefore, Fisher trace reveals how strongly a sample acts on the model, but not whether the induced update points in a trustworthy direction.
This radial ambiguity is one reason magnitude-based criteria can still suppress informative hard clean samples.

To address this ambiguity, we introduce relative geometric conflict (RGC), a direction-aware reliability principle.
Instead of judging the observed-label update in isolation, RGC compares it with a reference update induced by an EMA teacher \citep{tarvainen2017mean,laine2017temporal}.
If the observed label is correct but the sample is hard, its update can be large while remaining directionally aligned with the reference.
If the observed label is corrupted, the update is more likely to conflict with the reference direction.
We quantify this inconsistency by
$C_i=1-\cos(\mathbf{g}_i^{\mathrm{obs}},\mathbf{g}_i^{\mathrm{ref}})$,
which turns sample reliability estimation into a relative test of update consistency rather than an absolute test of difficulty or magnitude.
Our contributions are:
\begin{itemize}
    \item We reformulate noisy-label reliability estimation as the problem of assessing the reliability of observed-label-induced parameter updates, rather than measuring forward prediction difficulty.
    \item We use the sample-wise empirical Fisher trace as a backward-space diagnostic of update energy and make explicit why magnitude-only reliability criteria remain ambiguous.
    \item We propose relative geometric conflict, which compares observed-label and reference update directions to distinguish conflicting noisy updates from large but aligned hard-clean updates.
\end{itemize}

\section{Related Work}

\paragraph{Learning with noisy labels.}
Learning with noisy labels has been addressed by loss correction and robust objectives \citep{patrini2017loss,goldberger2017training,arazo2019unsupervised,zhang2018gce}, memorization-based sample selection or exchange \citep{jiang2018mentornet,han2018coteaching,yu2019does}, and semi-supervised or mixture-based label refinement \citep{li2020dividemix,longremix,promix,zhang2025psscl}.
Despite their differences, many methods estimate reliability from forward-space cues such as loss, confidence, entropy, prediction history, or small-loss selection.
These cues are useful but primarily measure learning difficulty rather than label reliability, and may therefore confuse hard clean samples with corrupted samples, especially under instance-dependent or feature-dependent noise \citep{xia2020part,PLC}.

\paragraph{Representation and gradient geometry.}
Several methods improve noisy-label robustness by shaping representations rather than only correcting labels. Contrastive learning provides a general tool for representation separation \citep{chen2020simple,he2020momentum,supervised}, and noisy-label variants use selective contrastive objectives, uniform selection, or progressive selection to improve feature discriminability \citep{unicon,zhang2025psscl}. These methods use representation geometry through feature similarity, class compactness, or selection confidence, whereas RGC tests whether the observed-label update is directionally consistent with a reference update in parameter space.
Information geometry views the Fisher matrix as a local metric, and Fisher/Hessian spectra are widely used to study curvature, sharpness, optimization, and generalization \citep{amari1998natural,ghorbani2019hessian,jastrzebski2021catastrophic}, though empirical Fisher approximations require care \citep{kunstner2019limitations}. Our sample-wise empirical trace is a local rank-one convention only for the individual matrix $\mathbf{g}_i\mathbf{g}_i^\top$, whose trace equals $\|\mathbf{g}_i\|_2^2$; it is used as an update-energy diagnostic rather than as a true or population Fisher estimate \citep{horn2012matrix,roy2007effective}.

\paragraph{Teacher and reference targets.}
Teacher-based learning and self-ensembling average predictions or weights to provide stable targets \citep{laine2017temporal,tarvainen2017mean}, and are widely used in noisy-label learning for pseudo-labeling, label refinement, and consistency regularization \citep{reed2015bootstrapping,berthelot2019mixmatch,li2020dividemix}.
Here, the EMA teacher is not merely a replacement label source; it defines a reference gradient for testing the geometric reliability of the observed-label update, and we explicitly treat teacher drift as a boundary condition.

\begin{figure}[t]
\centering
\includegraphics[width=1.0\linewidth]{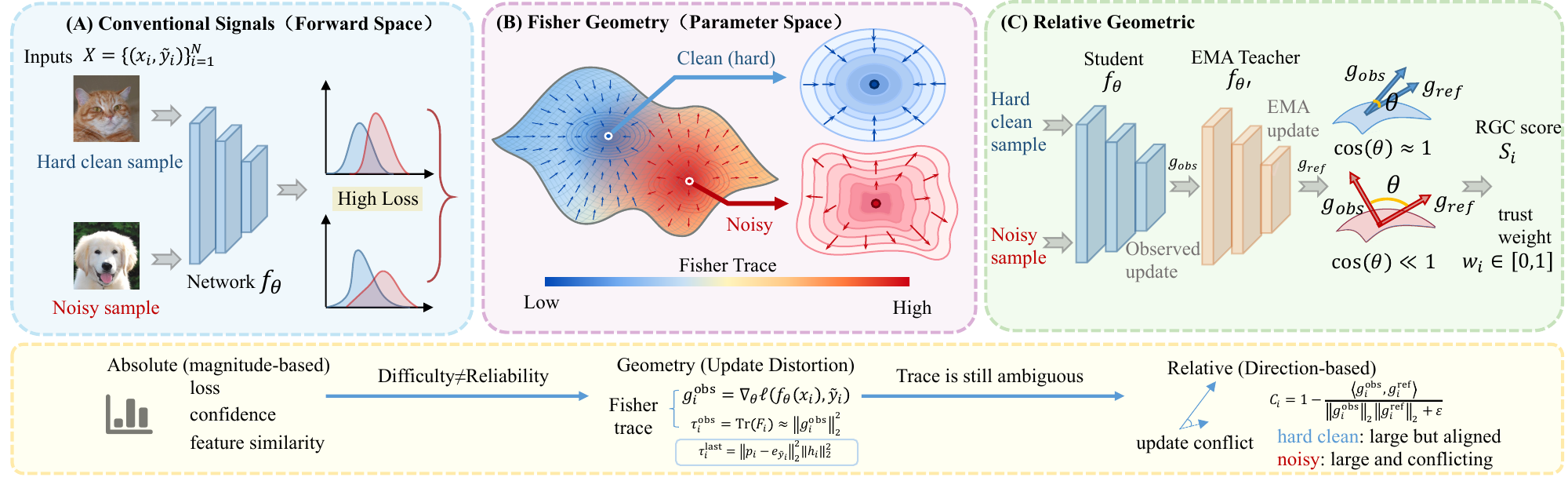}
\caption{
Overall motivation of RGC. Conventional forward-space signals mainly measure sample difficulty and confuse hard clean samples with noisy samples. Fisher trace moves the signal into parameter space but remains magnitude-based. RGC further compares update directions and helps identify unreliable samples by their directional conflict with a reference update.
}
\label{fig:1}
\end{figure}

\section{Method}
We propose a geometry-aware noisy-label learning framework that estimates whether the update induced by an observed label is reliable. The method has three ingredients: a temporally smoothed EMA reference target, a Fisher-trace proxy for update magnitude, and relative geometric conflict (RGC) for update direction. Magnitude measures how strongly a sample updates the model, while direction tests whether this update agrees with the reference.

\subsection{Problem Setup and Model Definition}
\label{sec:problem_setup}

Given a noisy training set
$\mathcal{D}=\{(x_i,\tilde{y}_i)\}_{i=1}^{N}$, the observed label
$\tilde{y}_i\in\{1,\ldots,K\}$ may differ from the latent clean label $y_i$.
We use a student network $f_\theta$ as the main model to be optimized, where
$\theta=(\psi,W)$ consists of a representation encoder $\phi_\psi$ and a classifier head $W$.
For each input $x_i$, the encoder extracts a representation
$h_i=\phi_\psi(x_i)$, and the classifier produces logits and class probabilities by
\[
z_i=Wh_i,
\qquad
p_i=\operatorname{softmax}(z_i).
\]

Our framework contains four functional modules.
First, the \emph{student prediction module} produces the observed-label update induced by the noisy annotation $\tilde{y}_i$.
Second, the \emph{EMA reference module} maintains a temporally smoothed teacher model, which provides the reference target for measuring whether the observed-label update is reliable.
Third, the \emph{geometric reliability module} computes sample-wise update statistics, including a lightweight Fisher-trace proxy for update magnitude and relative geometric conflict for update direction.
Finally, the \emph{trust-weighted optimization module} converts the reliability score into a sample weight $w_i\in[0,1]$, so that high-trust samples are weighted toward their observed labels and low-trust samples toward the reference target.

The reference target is produced by an EMA teacher $f_{\theta'}$ with the same architecture as the student:
\begin{align}
q_i
&=
\operatorname{Sharpen}\!\left(\operatorname{softmax}(f_{\theta'}(x_i));T\right),
\label{eq:teacher_target}
\\
\theta'_t
&\leftarrow
m\theta'_{t-1}+(1-m)\theta_t .
\label{eq:ema}
\end{align}
Here, $T$ denotes the sharpening temperature and $m$ is the EMA momentum.
The teacher is not assumed to be an oracle.
Instead, it acts as a lightweight temporal ensemble of historical student models, providing a smoother reference update without training multiple independent networks.
Compared with explicit ensemble or prototype-based references, the EMA teacher introduces one additional forward pass and keeps the extra computation limited.

\paragraph{Practical trace proxy.}
The geometric reliability module requires a sample-wise trace signal to measure the update magnitude induced by each example.
Throughout the paper, ``sample-wise Fisher trace'' means the empirical rank-one matrix formed from an individual sample gradient, $\mathbf{F}_i^{\mathrm{emp}}=\mathbf{g}_i\mathbf{g}_i^\top$.
Its trace is exactly $\operatorname{Tr}(\mathbf{F}_i^{\mathrm{emp}})=\|\mathbf{g}_i\|_2^2$; we do not use it as a population Fisher estimate.
For cross-entropy loss with observed label $\tilde{y}_i$, denote
\[
\delta_i = p_i-e_{\tilde{y}_i}.
\]
The full-model sample-wise empirical Fisher trace is the squared gradient norm over all trainable parameters:
\begin{equation}
\tau_i^{\mathrm{full}}
=
\|\nabla_\theta \ell_i\|_2^2
=
\|\nabla_W \ell_i\|_F^2
+
\|\nabla_\psi \ell_i\|_2^2 .
\label{eq:main_full_trace_decomp}
\end{equation}
Computing Eq.~\ref{eq:main_full_trace_decomp} for every sample is expensive.
We therefore use the classifier-layer contribution as a lightweight surrogate.
Since
\begin{equation}
\nabla_W \ell_i
=
\delta_i h_i^\top,
\end{equation}
the last-layer trace admits the closed form
\begin{equation}
\tau_i^{\mathrm{last}}
=
\|\nabla_W \ell_i\|_F^2
=
\|\delta_i h_i^\top\|_F^2
=
\|\delta_i\|_2^2\|h_i\|_2^2 .
\label{eq:main_last_trace}
\end{equation}
This surrogate avoids full per-sample backpropagation through the encoder and can be computed directly from the prediction residual and the feature norm.

Importantly, $\tau_i^{\mathrm{last}}$ is not an unconditional replacement for the full trace.
The approximation error is exactly the omitted representation-gradient energy:
\begin{equation}
\tau_i^{\mathrm{full}}
=
\tau_i^{\mathrm{last}}
+
\epsilon_i,
\qquad
\epsilon_i
=
\|\nabla_\psi \ell_i\|_2^2
\ge 0.
\label{eq:main_last_trace_residual}
\end{equation}
Thus, the last-layer trace is a lower-bound surrogate whose fidelity depends on whether $\epsilon_i$ is small relative to the classifier-layer contribution and to the trace margin between samples.
In particular, if $\epsilon_i\le \kappa_i\tau_i^{\mathrm{last}}$, then
\begin{equation}
\tau_i^{\mathrm{last}}
\le
\tau_i^{\mathrm{full}}
\le
(1+\kappa_i)\tau_i^{\mathrm{last}}.
\label{eq:main_last_trace_bound}
\end{equation}
\paragraph{Proposition 1.}
For samples $i$ and $j$, suppose $\epsilon_k\le \kappa_k\tau_k^{\mathrm{last}}$ for $k\in\{i,j\}$.
If
\begin{equation}
\tau_i^{\mathrm{last}}
>
(1+\kappa_j)\tau_j^{\mathrm{last}},
\label{eq:main_ordering_condition}
\end{equation}
then $\tau_i^{\mathrm{full}}>\tau_j^{\mathrm{full}}$.
This is a sufficient condition, not a necessary one: smaller last-layer gaps may still preserve the full-trace ordering when the omitted residuals are benign.


\subsection{From Forward Difficulty to Trace Geometry}

Forward-space criteria such as loss, confidence, or entropy are widely used to estimate sample reliability, but they primarily measure learning difficulty rather than label correctness.
A hard clean sample and a mislabeled sample can both produce high observed-label loss or low confidence, making them difficult to distinguish in prediction space.
To move beyond such forward-space ambiguity, we inspect the update induced by each training example:
\begin{equation}
\mathbf{g}_i^{\mathrm{obs}}
=
\nabla_\theta \ell(f_\theta(x_i),\tilde{y}_i).
\label{eq:g_obs}
\end{equation}
The sample-wise empirical Fisher trace provides a parameter-space measure of this update magnitude:
\begin{equation}
\mathbf{F}_i^{\mathrm{emp}}
=
\mathbf{g}_i^{\mathrm{obs}}(\mathbf{g}_i^{\mathrm{obs}})^\top,
\qquad
\tau_i^{\mathrm{obs}}
=
\operatorname{Tr}(\mathbf{F}_i^{\mathrm{emp}})
=
\|\mathbf{g}_i^{\mathrm{obs}}\|_2^2 .
\label{eq:trace_approx}
\end{equation}
This equality is an empirical rank-one convention for the individual sample gradient, not a claim about estimating the population Fisher.
This trace signal captures how strongly a sample acts on the model parameters, and therefore provides information that is not visible from loss or confidence alone.
In practice, we use the classifier-layer surrogate derived in Sec.~\ref{sec:problem_setup}.

However, trace remains a magnitude-based criterion.
A hard clean sample may induce a large but useful update, while a noisy sample may induce a large and misleading update.
Thus, update strength alone cannot determine whether the observed-label supervision is reliable.
This motivates the next step: comparing the \emph{direction} of the observed-label update with a reference update, which leads to relative geometric conflict.

\subsection{Relative Geometric Conflict}

RGC addresses the remaining ambiguity by comparing the observed-label update with a reference update.
To remain consistent with the last-layer trace surrogate above, we compute this comparison on the classifier head $W$ rather than on the full parameter set $\theta$.
Given $q_i$ from Eq.~\ref{eq:teacher_target}, define
\begin{equation}
\mathbf{g}_{i,W}^{\mathrm{obs}}
=
\nabla_W \ell(f_\theta(x_i),\tilde y_i),
\qquad
\mathbf{g}_{i,W}^{\mathrm{ref}}
=
\nabla_W \ell(f_\theta(x_i),q_i),
\qquad
C_i^\epsilon
=
1-
\frac{
\langle \mathbf{g}_{i,W}^{\mathrm{obs}},\mathbf{g}_{i,W}^{\mathrm{ref}}\rangle
}{
\|\mathbf{g}_{i,W}^{\mathrm{obs}}\|_F
\|\mathbf{g}_{i,W}^{\mathrm{ref}}\|_F+\epsilon
}.
\label{eq:conflict}
\end{equation}
Small $C_i^\epsilon$ means that the observed-label update is aligned with the reference; large $C_i^\epsilon$ indicates directional conflict. This distinction targets the case trace alone can mix: hard clean samples can have large gradient norm but small conflict, whereas mislabeled samples tend to induce directions that disagree with the reference.

For the softmax classifier, these gradients admit the closed forms
\begin{equation}
\mathbf{g}_{i,W}^{\mathrm{obs}}
=
(p_i-e_{\tilde y_i})h_i^\top,
\qquad
\mathbf{g}_{i,W}^{\mathrm{ref}}
=
(p_i-q_i)h_i^\top .
\label{eq:conflict_last_layer_forms}
\end{equation}
The score uses $\tau_i^{\mathrm{obs}}=\|p_i-e_{\tilde y_i}\|_2^2\|h_i\|_2^2$ and $\tau_i^{\mathrm{ref}}=\|p_i-q_i\|_2^2\|h_i\|_2^2$; when the feature and residual norms are nonzero, $C_i^\epsilon$ is the stabilized implementation of the resulting logit-residual angular discrepancy.
We use classifier-space conflict for the same reason we use last-layer trace: it is the local interface where the observed label acts on the representation, providing a lightweight proxy for whether supervision is geometrically consistent at the decision boundary.

The discrepancy between the EMA target and the latent clean target controls an upper bound on the reference-gradient error. Let $q_i^\star$ be the ideal clean target and assume the logit-to-parameter Jacobian satisfies $\|J_i\|_2\le L_i$. For soft-label cross-entropy,
\begin{equation}
\left\|
\nabla_\theta \ell(p_\theta(x_i), q_i)
-
\nabla_\theta \ell(p_\theta(x_i), q_i^\star)
\right\|_2
\le
L_i \|q_i-q_i^\star\|_2 .
\label{eq:teacher_gradient_error}
\end{equation}
When the teacher drifts toward noisy labels, the target discrepancy can grow and the reference can become harmful. In practice, we mitigate this failure mode with warm-up, per-sample score smoothing, and a gated objective that applies teacher supervision mainly to low-trust samples.

This observed--reference comparison also specifies the roles of trace and RGC.
For the nondegenerate theoretical case with $a_i=\|\mathbf{g}_{i,W}^{\mathrm{obs}}\|_F>0$ and $b_i=\|\mathbf{g}_{i,W}^{\mathrm{ref}}\|_F>0$, let $r_i=a_i/b_i$ and define the unstabilized angular conflict $C_i^\circ=1-\cos\theta_i$.
The normalized gradient discrepancy decomposes as
\begin{equation}
\frac{
\|\mathbf{g}_{i,W}^{\mathrm{obs}}-\mathbf{g}_{i,W}^{\mathrm{ref}}\|_F^2
}{b_i^2}
=
(r_i-1)^2+2r_iC_i^\circ .
\label{eq:gradient_discrepancy_decomp}
\end{equation}
This identity is exact for the unstabilized geometry; trace captures radial mismatch, while RGC uses $C_i^\epsilon$ as a numerically robust angular signal.
This decomposition reframes the score from update strength alone to whether the observed-label update is geometrically consistent with the reference.

\subsection{Reliability Score and Overall Objective}
\label{sec:optimization_objective}


We convert the observed--reference update discrepancy into a sample-level unreliability score.
Given the observed-label trace $\tau_i^{\mathrm{obs}}$, the reference trace $\tau_i^{\mathrm{ref}}$, and the stabilized relative geometric conflict $C_i^\epsilon$ in Eq.~\ref{eq:conflict}, we define
\begin{equation}
R_i^\tau
=
\log\frac{\tau_i^{\mathrm{obs}}+\epsilon}
{\tau_i^{\mathrm{ref}}+\epsilon},
\qquad
s_i
=
\beta\widehat{R}_i^\tau+(1-\beta)\widehat{C}_i^\epsilon ,
\label{eq:score}
\end{equation}
where $\widehat{R}_i^\tau$ and $\widehat{C}_i^\epsilon$ are standardized within each fine-tuning epoch.
Here, $R_i^\tau$ measures radial mismatch in update magnitude, while $C_i^\epsilon$ measures angular mismatch in update direction.
We use a direction-dominant default because trace can be large for both noisy samples and hard clean samples; directional conflict is meant to test agreement with the reference rather than provide an optimal mixing rule.

\paragraph{Proposition 2.}
For a noisy sample $n$ and a hard clean sample $h$, define
\[
\Delta_C=\widehat{C}_n-\widehat{C}_h,
\qquad
\Delta_R=\widehat{R}_n^\tau-\widehat{R}_h^\tau .
\]
If the directional margin satisfies $\Delta_C\ge m_C>0$ and the radial term is no worse than $\Delta_R\ge -m_R$ for $m_R\ge 0$, then the fused score preserves the noisy-over-hard-clean ordering, $s_n>s_h$, for any
\begin{equation}
0
\le
\beta
<
\frac{m_C}{m_C+m_R}.
\label{eq:main_beta_interval}
\end{equation}
Indeed, $s_n-s_h\ge (1-\beta)m_C-\beta m_R>0$, so this sufficient interval keeps the angular term dominant when radial trace alone is ambiguous.

The raw unreliability score is smoothed by a per-sample EMA and mapped to the observed-label trust weight:
\begin{equation}
\tilde{s}_{i,t}
\leftarrow
\mu \tilde{s}_{i,t-1}
+
(1-\mu)s_{i,t},
\qquad
w_i
=
\sigma\!\left(
-\alpha
\frac{\tilde{s}_{i,t}-\mathrm{mean}(\tilde{s}_{t})}
{\mathrm{std}(\tilde{s}_{t})+\epsilon}
\right).
\label{eq:weight}
\end{equation}
A larger $s_i$ yields a smaller $w_i$, indicating lower trust in the observed label.

The student is trained with a trust-weighted objective:
\begin{equation}
\mathcal{L}_{\mathrm{RGC}}
=
\frac{1}{N}\sum_{i=1}^{N}
\left[
w_i\ell(f_\theta(x_i),\tilde{y}_i)
+
\lambda(1-w_i)\ell(f_\theta(x_i),q_i)
\right].
\label{eq:final_objective}
\end{equation}
Thus, high-trust samples remain supervised mainly by their observed labels, while low-trust samples receive more reference-target supervision.
This continuous gate avoids a hard clean/noisy split: trust in the observed label decreases as its induced update becomes less consistent with the reference.

\begin{figure}[t]
    \centering
    \includegraphics[width=0.93\linewidth]{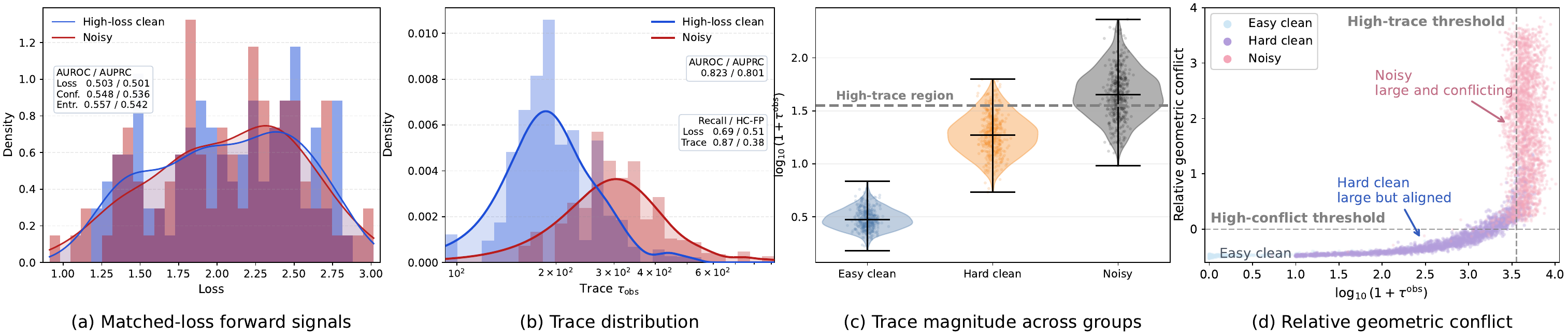}
    \caption{
Diagnostic analysis of reliability signals.
Under matched difficulty, forward-space signals become non-discriminative; trace improves detection but remains vulnerable to hard-clean false positives; relative geometric conflict further separates hard clean and noisy samples along the directional dimension.
}
    \label{fig:diagnostic_signals}
\end{figure}

\section{Experiments}
\label{sec:experiments}

\paragraph{Setup.}
We evaluate Trace and RGC on CIFAR-10 and CIFAR-100 under symmetric and asymmetric noise, and on Animal-10N and Food-101N under real-world label noise.
CIFAR noise rates range from 20\% to 80\%.
Unless otherwise specified, the baseline includes warm-up, instance-level difficulty modeling, and dynamic optimization \citep{IDO}.
RGC adds three reliability modules: an EMA teacher that provides the reference target, a trace-based radial reliability signal, and the relative geometric conflict term for directional reliability.
All entries are evaluated under a unified 30-epoch realistic fine-tuning protocol.
Concretely, the model is first warmed up for $E_w=5$ epochs using observed-label cross-entropy and then trained with an EMA teacher ($m=0.999$), sharpened targets ($T=0.7$), score smoothing ($\mu=0.9$), trust sharpness $\alpha=0.6$, and radial--angular fusion coefficient $\beta=0.2$.
We report classification accuracy, noisy-sample detection, and hard-clean preservation.
For diagnostic evaluation on synthetic-noise benchmarks, noisy samples are $\mathcal{N}=\{i:\tilde{y}_i\neq y_i\}$, and hard clean samples $\mathcal{H}$ are defined as the top-20\% high-loss subset among clean samples.
Given a threshold $\eta$, samples with $s_i>\eta$ are predicted as unreliable, and we report
\[
\mathrm{NoisyRec.}
=
\frac{|\{i\in\mathcal{N}:s_i>\eta\}|}{|\mathcal{N}|},
\quad
\mathrm{HC\mbox{-}FP}
=
\frac{|\{i\in\mathcal{H}:s_i>\eta\}|}{|\mathcal{H}|},
\quad
\mathrm{HCPR}
=
1-\mathrm{HC\mbox{-}FP}.
\]
These metrics test whether a reliability signal detects corrupted labels while preserving hard but correctly labeled examples.

\subsection{Diagnostic analysis of reliability signals.}
\label{sec:diagnostic_compact}

We construct a loss-overlapped diagnostic subset by selecting the top 20\% clean samples with the highest loss after warm-up as hard-clean samples and then comparing them with noisy samples. As shown in Fig.~\ref{fig:diagnostic_signals}(a), hard clean and noisy samples have nearly overlapping observed-label loss distributions, so forward-space signals are near chance: loss achieves only 0.503/0.501 AUROC/AUPRC, while confidence and entropy remain close to chance. Fig.~\ref{fig:diagnostic_signals}(b) shows that trace partially reduces this ambiguity by moving the reliability signal into parameter space, reaching 0.823/0.801 AUROC/AUPRC and indicating that update magnitude contains information beyond scalar loss.

However, Fig.~\ref{fig:diagnostic_signals}(c) shows that trace remains a radial signal: hard clean samples can still overlap with noisy samples in the high-trace region. Fig.~\ref{fig:diagnostic_signals}(d) shows that adding relative geometric conflict helps separate these cases along the conflict dimension: hard clean samples may still induce large updates, but they remain comparatively aligned with the reference, whereas noisy samples tend to occupy the high-conflict region. Together, these diagnostics are consistent with the intended use of RGC: forward-space signals are weak under matched difficulty, trace adds backward-space magnitude, and relative conflict helps distinguish large useful updates from large harmful updates.


\subsection{Main Results}

Tables~\ref{tab:cifar10_results} and~\ref{tab:realworld_results} report the main comparison results.
Under this evaluation protocol, RGC is the top entry among the evaluated methods in each CIFAR-10 noise setting.
The largest gaps appear in the 80\% symmetric and 40\% asymmetric settings, with improvements of +6.2 and +3.3 points over the next-highest non-RGC entries in those columns.
This pattern is consistent with the intended role of RGC. In these regimes, forward-space cues become less reliable because hard clean and corrupted samples can both appear difficult. Trace partially improves reliability estimation by measuring update magnitude, but hard clean samples near class boundaries can also induce large informative gradients, so magnitude alone still mixes large useful updates with large harmful ones. RGC adds a directional test relative to the EMA reference: hard clean samples can remain large but aligned, whereas corrupted samples are more likely to deviate from the reference direction. The mechanism is evaluated directly in Sec.~\ref{sec:diagnostic_compact} and in the component ablations below.

\begin{table}[tb]
    \caption{
    Test accuracy (\%) on the CIFAR-10 and CIFAR-100 datasets under symmetric and asymmetric noise.
    Bold indicates the highest value in each column.
    }
    \label{tab:cifar10_results}
    \centering
    \scriptsize
    \setlength{\tabcolsep}{3pt}
    \begin{tabularx}{\textwidth}{l *{9}{Y}}
        \toprule
        \multirow{2}{*}{Method}
        & \multicolumn{5}{c}{CIFAR-10}
        & \multicolumn{4}{c}{CIFAR-100} \\
        \cmidrule(lr){2-6} \cmidrule(lr){7-10}
        & \multicolumn{3}{c}{Sym.}
        & Asym.
        & Avg.
        & \multicolumn{2}{c}{Sym.}
        & Asym.
        & Avg. \\
        \cmidrule(lr){2-4}
        \cmidrule(lr){5-5}
        \cmidrule(lr){7-8}
        \cmidrule(lr){9-9}
        & 20\% & 50\% & 80\% & 40\% &
        & 20\% & 50\% & 40\% & \\
        \midrule
        Standard CE
        & 73.4 & 58.9 & 49.2 & 60.9 & 60.6
        & 44.6 & 32.8 & 38.7 & 38.7 \\

        GCE \cite{zhang2018gce}
        & 77.2 & 75.7 & 66.7 & 67.7 & 71.8
        & 50.1 & 42.3 & 45.7 & 46.0 \\

        DivideMix \cite{li2020dividemix}
        & 79.5 & 74.6 & 70.7 & 68.1 & 73.2
        & 61.6 & 47.9 & 53.1 & 54.2 \\

        RRL \cite{rrl}
        & 88.4 & 83.3 & 74.7 & 76.5 & 80.7
        & 61.0 & 58.6 & 53.5 & 57.7 \\

        UNICON \cite{unicon}
        & 72.8 & 71.5 & 68.1 & 76.4 & 72.2
        & 60.7 & 57.1 & 54.2 & 57.3 \\

        ProMix \cite{promix}
        & 87.4 & 83.7 & 68.7 & 77.0 & 79.2
        & 55.2 & 52.8 & 44.6 & 50.9 \\
        LongReMix \cite{longremix}
        & 81.4 & 79.0 & 67.6 & 67.7 & 73.9
        & 61.3 & 56.9 & 47.8 & 55.3 \\

        L2B \cite{l2b}
        & 86.7 & 82.1 & 59.4 & 76.7 & 76.2
        & 42.9 & 46.1 & 39.8 & 42.9 \\

        PSSCL \cite{zhang2025psscl}
        & 82.3 & 80.7 & 77.6 & 71.0 & 77.9
        & 56.5 & 45.6 & 46.0 & 49.4 \\

        DCD \cite{dcd}
        & 87.6 & 78.7 & 72.9 & 75.8 & 78.8
        & 63.3 & 54.9 & 53.3 & 57.2 \\

        EarlyCut \cite{mees}
        & 87.9 & 83.7 & 78.4 & 76.6 & 81.7
        & 63.7 & 59.6 & 52.0 & 58.4 \\

        IDO \cite{IDO}
        & 90.4 & 86.9 & 80.7 & 84.3 & 85.6
        & 70.8 & 63.2 & 60.5 & 64.8 \\

        \textbf{RGC (Ours)}
        & \textbf{93.5} & \textbf{92.4} & \textbf{86.9} & \textbf{87.6} & \textbf{90.1}
        & \textbf{73.7} & \textbf{68.6} & \textbf{64.6} & \textbf{69.0} \\
        \bottomrule
    \end{tabularx}
\end{table}

\begin{figure}[t]
\centering
\begin{minipage}[t]{0.48\linewidth}
    \centering
    \vspace{0pt}
    \captionof{table}{
    Test accuracy (\%) on real-world noisy-label datasets.
    Bold indicates the highest value in each column.
    }
    \label{tab:realworld_results}
    \scriptsize
    \setlength{\tabcolsep}{3pt}
    \begin{tabularx}{\linewidth}{lYcc}
        \toprule
        Method & Ref. & Animal-10N & Food-101N\\
        \midrule
        DivideMix & ICLR20 & 71.5 & 67.2\\
        RRL       & ICCV21 & 73.8 & 69.7\\
        UNICON    & CVPR22 & 71.1 & 70.4\\
        ProMix    & IJCAI23 & 76.2 & 70.1\\
        LongReMix & PR23    & 71.6 & 68.7\\
        L2B       & CVPR24  & 71.4 & 68.9\\
        PSSCL     & PR25    & 72.7 & 70.7\\
        DCD       & ICCV25  & 74.1 & 71.2\\
        EarlyCut  & NeurIPS25 & 75.0 & 72.1\\
        IDO       & NeurIPS25 & 77.2 & 74.4\\
        \midrule
        \textbf{RGC (Ours)} & -- & \textbf{81.4} & \textbf{78.5}\\
        \bottomrule
    \end{tabularx}
\end{minipage}
\hfill
\begin{minipage}[t]{0.48\linewidth}
    \centering
    \vspace{0pt}
    \includegraphics[width=0.90\linewidth]{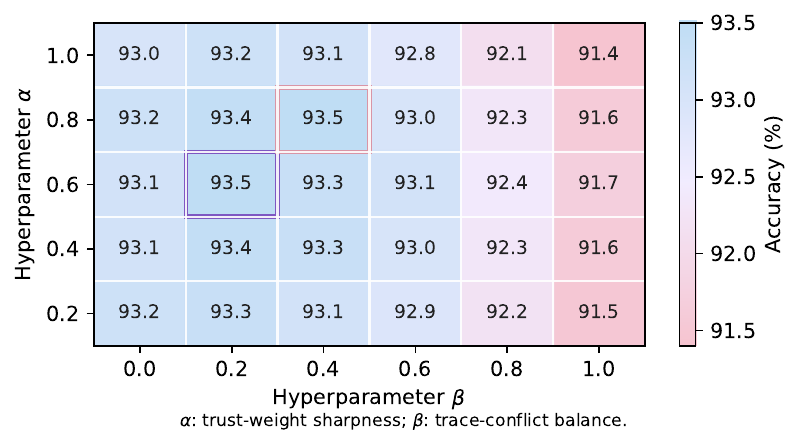}
    \captionof{figure}{
    Hyperparameter sensitivity of RGC with respect to $\alpha$ and $\beta$.
    $\alpha$ controls trust-weight sharpness, and $\beta$ controls the balance between radial trace magnitude and directional conflict.
    }
    \label{fig:alpha_beta}
\end{minipage}
\end{figure}

\subsection{Ablation Studies}

\subsubsection{Ablation on Geometric Reliability Components}
Table~\ref{tab:component_ablation} studies the three components of geometric reliability.
Here, Teacher denotes the same underlying noisy-label training pipeline augmented with EMA-based reference supervision, but without trace-based reliability scoring or conflict modeling.
Trace-only uses absolute last-layer trace, Trace+Teacher uses relative radial mismatch, and full RGC adds conflict.
Starting from the trace-only variant, adding Teacher improves most settings but is less stable in some cases, suggesting that teacher supervision by itself can still drift.
Combining Trace and Teacher improves the row further, indicating that magnitude information and reference supervision are complementary.
With Conflict added, the full RGC model has the highest or tied-highest entries in this ablation, suggesting that directional inconsistency provides additional reliability information beyond magnitude alone.

\begin{table}[tb]
\centering
\caption{
Ablation on the geometric reliability components on CIFAR-10 and CIFAR-100.
Trace denotes radial update-magnitude scoring, Teacher denotes the EMA-based reference target, and Conflict denotes the proposed relative geometric conflict.
}
\label{tab:component_ablation}
\scriptsize
\setlength{\tabcolsep}{3pt}
\begin{tabularx}{\textwidth}{lccc *{7}{Y}}
\toprule
\multirow{2}{*}{Variant}
& \multicolumn{3}{c}{Module}
& \multicolumn{4}{c}{CIFAR-10}
& \multicolumn{3}{c}{CIFAR-100} \\
\cmidrule(lr){2-4}
\cmidrule(lr){5-8}
\cmidrule(lr){9-11}
& Trace
& Teacher
& Conflict
& Sym. 20\%
& Sym. 50\%
& Sym. 80\%
& Asym. 40\%
& Sym. 20\%
& Sym. 50\%
& Asym. 40\% \\
\midrule
Trace
& \cmark & \xmark & \xmark
& 91.5 & 88.4 & 83.0 & 85.8
& 71.2 & 64.4 & 61.7 \\

Teacher
& \xmark & \cmark & \xmark
& 91.2 & 88.7 & 82.5 & 85.4
& 71.4 & 64.9 & 62.5 \\

Trace+Teacher
& \cmark & \cmark & \xmark
& 92.6 & 90.8 & 85.1 & 86.8
& 72.2 & 66.8 & 64.4 \\

\textbf{RGC (Ours)}
& \cmark & \cmark & \cmark
& \textbf{93.5} & \textbf{92.4} & \textbf{86.9} & \textbf{87.6}
& \textbf{73.7} & \textbf{68.6} & \textbf{64.6}  \\
\bottomrule
\end{tabularx}
\end{table}

\begin{figure}[t]
    \centering
    \includegraphics[width=\linewidth]{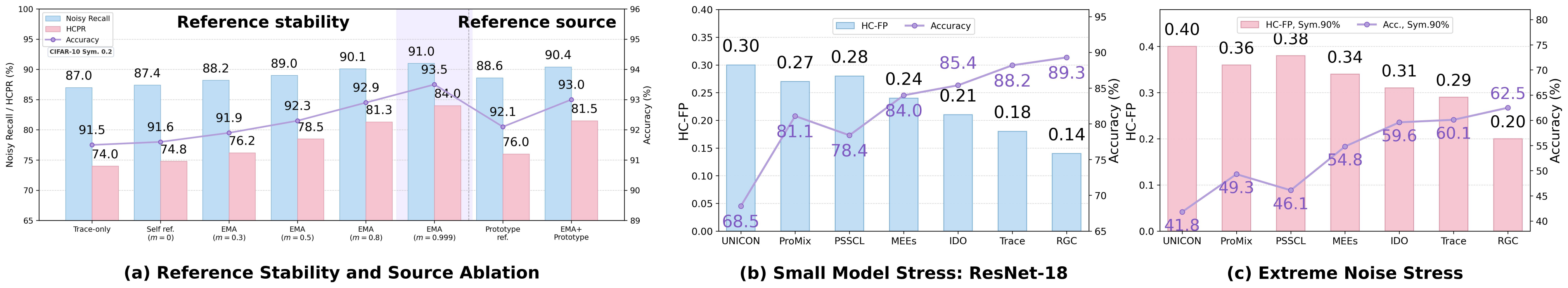}
    \caption{
Ablation and boundary analysis of RGC.
The figure summarizes the reference design ablation and the stress tests under weak model capacity and extreme noise.
}
    \label{fig:ablation_row3}
\end{figure}

\subsubsection{Hyperparameter Sensitivity}

We study the sensitivity to $\alpha$ and $\beta$, where $\alpha$ controls the trust-weight sharpness and $\beta$ balances radial trace magnitude with angular conflict.
As shown in Fig.~\ref{fig:alpha_beta}, RGC changes little over the tested range of $\alpha$.
This behavior is plausible because $\alpha$ only controls how aggressively the unreliability score is mapped to the observed-label trust weight; it does not change the underlying reliability ordering of samples.
In contrast, $\beta$ directly changes the reliability signal itself.
When $\beta$ is small to moderate, directional conflict dominates and trace serves as an auxiliary magnitude cue, which is consistent with preserving hard clean samples with large but aligned updates while down-weighting corrupted samples with conflicting updates.
When $\beta$ becomes large, the score is dominated by trace magnitude and moves toward a magnitude-only criterion, which again confuses useful hard-clean updates with harmful noisy updates.
The performance drop at large $\beta$ is consistent with the view that the key signal in RGC is not update magnitude alone, but the relative direction between the observed-label update and the EMA reference update.

\subsubsection{Ablation on Teacher}

The reference ablation indicates that the quality of the reference direction matters. Trace-only has no teacher and can only measure radial update magnitude, so it still mixes large harmful noisy updates with large useful hard-clean updates. When $m=0$, the teacher collapses to a self-reference that is tightly coupled with noisy mini-batch updates and only slightly improves over Trace-only. As $m$ increases, the EMA teacher follows $\theta'_t = m\theta'_{t-1} + (1-m)\theta_t$ and becomes a temporally smoothed model, which suppresses short-term fluctuations and yields a more stable reference direction in these experiments. The resulting accuracy and HCPR pattern is consistent with the intended role of the reference: preserve hard clean samples with large but aligned updates and down-weight corrupted samples with conflicting directions. Prototype-based references are less stable because class prototypes are computed from evolving noisy representations and may drift toward corrupted regions; combining prototypes with EMA therefore does not consistently outperform pure EMA. These results support using EMA as a reference direction rather than as an oracle pseudo-labeler.

\subsection{Boundary and Robustness Validation}

We evaluate RGC under low-data, small-backbone, and extreme-noise settings to examine when the reference direction becomes unreliable and how this limits the gains of direction-aware reliability estimation.

\paragraph{Low-Data Boundary}
Table~\ref{tab:low_data_reliability} shows that all methods degrade as the amount of training data decreases, while HC-FP generally rises, confirming that low-data training weakens reliability estimation.
RGC remains above IDO across all data ratios, but the hard-clean false-positive rate worsens as the data ratio decreases.
At the extreme 20\% setting, Trace slightly surpasses RGC in accuracy (84.7 vs.~84.3) and HC-FP (0.46 vs.~0.49), while RGC remains above IDO in accuracy. This marks a boundary case: with very little data, the EMA reference becomes higher-variance, so conflict may no longer distinguish useful hard-clean updates from harmful noisy ones.

\begin{table}[tb]
\centering
\caption{
Low-data robustness under different training-data ratios.
Higher Acc. and lower HC-FP are better; $\Delta=\mathrm{RGC}-\mathrm{IDO}$.
}
\label{tab:low_data_reliability}
\scriptsize
\setlength{\tabcolsep}{3pt}
\begin{tabular}{llccccccc}
\toprule
Data Ratio & Metric
& ProMix & PSSCL & EarlyCut & IDO & Trace & \textbf{RGC} & $\Delta$ \\
\midrule
\multirow{2}{*}{100\%}
& Acc. $\uparrow$
& 87.4 & 82.3 & 87.9 & 90.4 & 91.5 & \textbf{93.5} & +3.1 \\
& HC-FP $\downarrow$
& 0.34 & 0.32 & 0.29 & 0.25 & 0.22 & \textbf{0.16} & -0.09 \\
\midrule
\multirow{2}{*}{80\%}
& Acc. $\uparrow$
& 86.1 & 80.9 & 86.4 & 88.7 & 90.7 & \textbf{92.1} & +3.4 \\
& HC-FP $\downarrow$
& 0.36 & 0.39 & 0.31 & 0.32 & 0.27 & \textbf{0.23} & -0.09 \\
\midrule
\multirow{2}{*}{50\%}
& Acc. $\uparrow$
& 82.7 & 77.2 & 83.9 & 85.4 & 89.2 & \textbf{90.8} & +5.4 \\
& HC-FP $\downarrow$
& 0.46 & 0.47 & 0.40 & 0.38 & 0.34 & \textbf{0.28} & -0.10 \\
\midrule
\multirow{2}{*}{20\%}
& Acc. $\uparrow$
& 76.9 & 71.0 & 79.1 & 79.0 & \textbf{84.7} & 84.3 & +5.3 \\
& HC-FP $\downarrow$
& 0.58 & 0.60 & 0.54 & 0.51 & \textbf{0.46} & 0.49 & -0.02 \\
\bottomrule
\end{tabular}
\end{table}

\begin{figure}[t]
\centering

\begin{minipage}[t]{0.47\linewidth}
\centering
\vspace{0pt}
\captionof{table}{
Runtime and memory on CIFAR-10 with 20\% symmetric noise.
}
\label{tab:runtime_memory_representative}
\scriptsize
\setlength{\tabcolsep}{4pt}
\begin{tabular}{lcc}
\toprule
Method
& Sec./Epoch $\downarrow$
& Memory $\downarrow$ \\
\midrule
CE
& 52
& 3.2 GB \\

DivideMix
& 138
& 8.4 GB \\

PSSCL
& 165
& 8.6 GB \\

ProMix
& 121
& 6.3 GB \\

DCD
& 114
& 6.0 GB \\

IDO
& 98
& 4.9 GB \\

\textbf{RGC}
& 110
& 5.3 GB \\

\bottomrule
\end{tabular}
\end{minipage}
\hfill
\begin{minipage}[t]{0.47\linewidth}
\centering
\vspace{0pt}
\includegraphics[width=0.70\linewidth]{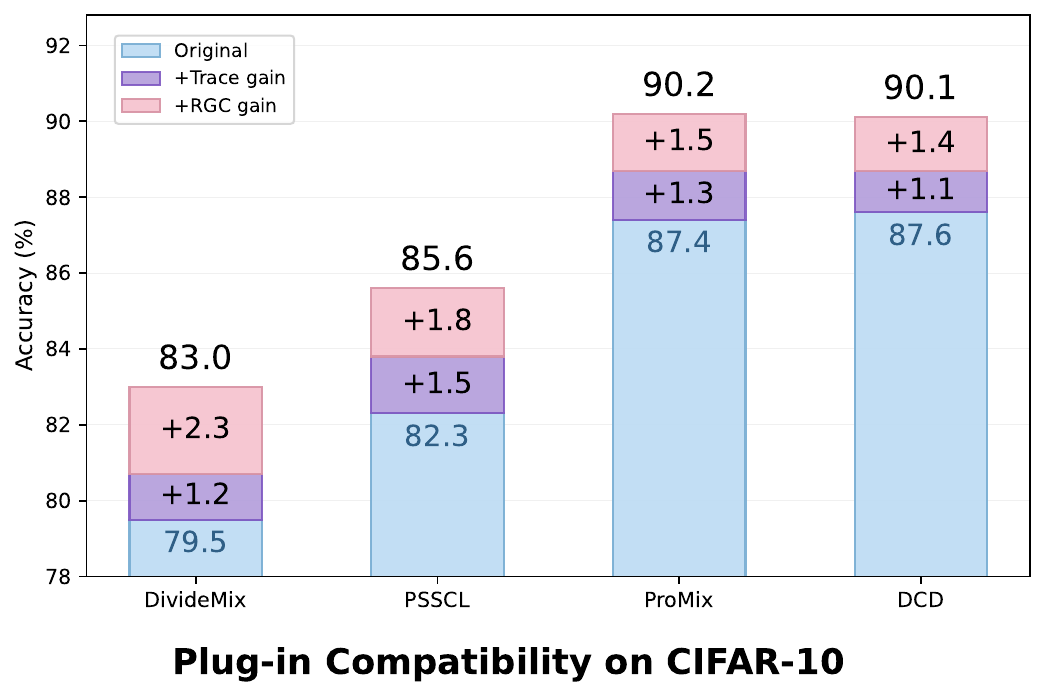}
\captionof{figure}{
Plug-in compatibility of Trace and RGC on representative noisy-label learning frameworks.
}
\label{fig:plugin_compatibility_c10}
\end{minipage}

\end{figure}

\paragraph{Model Capacity and Extreme Noise}
Model capacity and extreme noise stress the geometric reference. A smaller backbone weakens representations and reduces the gain over Trace. Extreme symmetric noise can drift the teacher toward corrupted labels, while extreme asymmetric noise can make wrong-class updates partly aligned; in both cases, RGC becomes more reference-limited.

\vspace{-0.4em}
\section{Efficiency and Plug-in Compatibility}
\label{sec:efficiency_plugin}

Table~\ref{tab:runtime_memory_representative} lists per-epoch runtime and peak GPU memory on CIFAR-10 with 20\% symmetric label noise. Cross-entropy is fastest but is not a noisy-label method. DivideMix, PSSCL, ProMix, and DCD land at 114--165\,s per epoch and 6.0--8.6\,GB, which tracks their co-training, contrastive batches, or extra objectives. IDO takes 98\,s and 4.9\,GB; RGC moves to 110\,s and 5.3\,GB. The gap is small next to those heavier pipelines and matches what Sec.~\ref{sec:problem_setup} adds: last-layer trace and conflict, not a full-model Fisher or an added network pass beyond the usual teacher setup.

Fig.~\ref{fig:plugin_compatibility_c10} keeps each host method's training recipe but swaps in Trace and then RGC for the reliability module. Trace uses parameter-space magnitude; RGC compares the observed-label update to the teacher reference direction. Under this setup, accuracy moves up in the figure while Table~\ref{tab:runtime_memory_representative} places RGC below DivideMix, PSSCL, ProMix, and DCD on time and memory, with a modest runtime reduction relative to DCD (110 versus 114\,s per epoch).

\vspace{-0.5em}
\section{Limitations}
RGC relies on the reference direction being more stable than the observed-label update, which can weaken when the EMA teacher has high variance or drifts toward corrupted labels. The last-layer trace and conflict proxies focus on the classifier interface rather than full-model geometry; the sufficient conditions discussed here do not provide a general guarantee for full-model geometry. The $\beta$ interval is likewise sufficient, not an optimal tuning rule. Larger architectures and non-classification tasks remain future work.

\vspace{-0.5em}
\section{Conclusion}

We presented RGC, a direction-aware reliability criterion for noisy-label learning that compares the observed-label update with a teacher-induced reference update. It combines radial update energy with observed--reference conflict to score trust in noisy supervision. In the evaluated protocols, RGC improves over magnitude-only trace criteria when the reference is reliable, with gains shrinking when the teacher can drift. Code is available at \url{https://anonymous.4open.science/r/AAA-BF68/}.

\bibliographystyle{plainnat}
\bibliography{references}


\end{document}